\documentclass[runningheads]{llncs}
\usepackage{graphicx}
\usepackage[usenames,dvipsnames]{xcolor}

\usepackage{pifont}
\newcommand{\cmark}{\ding{51}}%
\newcommand{\xmark}{\ding{55}}%
\usepackage{latexsym}
\usepackage{amssymb}
\usepackage{amsmath}
\usepackage{hyperref}
\usepackage{multirow}
\usepackage{booktabs}

\usepackage{ifsym}

\newcommand*{\defeq}{\mathrel{\vcenter{\baselineskip0.5ex \lineskiplimit0pt
			\hbox{\footnotesize.}\hbox{\footnotesize.}}}%
	=}
\newcommand{\avgRank}{\texttt{AvgRank}}
\newcommand{\randomRank}{\texttt{RandomRank}}
\newcommand{\avgPrf}{\texttt{AvgPerformance}}

\newcommand{\AlorsNDCG}{\texttt{Alors (NDCG)}}
\newcommand{\AlorsRegr}{\texttt{Alors (REGR)}}

\newcommand{\regretAtK}[1]{regret@$#1$}

\begin{document}
\title{Extreme Algorithm Selection with\\ Dyadic Feature Representation\thanks{This paper is published at the Discovery Science 2020 conference. The final authenticated version is available online at \url{https://doi.org/10.1007/978-3-030-61527-7_21}.}}
%
%
\author{Alexander Tornede\inst{1}\orcidID{0000-0002-2415-2186} \and
Marcel Wever\inst{1}\orcidID{0000-0001-9782-6818} \and
Eyke H{\"u}llermeier\inst{1}\orcidID{0000-0002-9944-4108}}
\authorrunning{A. Tornede et al.}
%
\institute{Heinz Nixdorf Institute and Department of Computer Science, Paderborn University, Warburger Str. 100, 33100 Paderborn, Germany\\
\email{\{alexander.tornede, marcel.wever, eyke\}@uni-paderborn.de}
}
\maketitle              
\begin{abstract}
Algorithm selection (AS) deals with the automatic selection of an algorithm from a fixed set of candidate algorithms most suitable for a specific instance of an algorithmic problem class, e.g., choosing solvers for SAT problems. Benchmark suites for AS usually comprise candidate sets consisting of at most tens of algorithms, whereas in algorithm configuration (AC) and combined algorithm selection and hyperparameter optimization (CASH) problems the number of candidates becomes intractable, impeding to learn effective meta-models and thus requiring costly online performance evaluations.
In this paper, we propose the setting of extreme algorithm selection (XAS), which, despite assuming limited time resources and hence excluding online evaluations at prediction time, allows for considering thousands of candidate algorithms and thereby facilitates meta learning. We assess the applicability of state-of-the-art AS techniques to the XAS setting and propose approaches leveraging a dyadic representation, in which both problem instances and algorithms are described in terms of feature vectors. We find this approach to significantly improve over the current state of the art in various metrics.

\end{abstract}
\section{Introduction}

Algorithm selection (AS) refers to a specific recommendation task, in which the choice alternatives are algorithms: Given a set of candidate algorithms to choose from, and a specific instance of a problem class, such as SAT or integer optimization, the task is to select or recommend an algorithm that appears to be most suitable for that instance, in the sense of performing best in terms of criteria such as runtime, solution quality, etc.
Hitherto practical applications of AS, as selecting a SAT solver for a logical formula, typically comprise candidate sets consisting of at most tens of algorithms, and this is also the order of magnitude that is found in standard AS benchmark suites such as ASlib \cite{bischl2016aslib}.

This is in contrast with the problem of combined algorithm selection and hyperparameter optimization (CASH) \cite{thornton2013auto} as considered in automated machine learning (AutoML), where the number of potential candidates is very large and potentially infinite \cite{thornton2013auto,feurer2015efficient,MohrWH18}.
Corresponding methods heavily rely on computationally extensive search procedures combined with costly online evaluations of the performance measure to optimize for, since learning effective meta models for an instantaneous recommendation becomes infeasible.

In this paper, we propose \emph{extreme algorithm selection} (XAS) as a novel setting in-between traditional AS and AC/CASH, which is motivated by application scenarios characterized by
\begin{itemize}
\item the demand for prompt recommendations in quasi real time, 
\item an extremely large (though still finite) set of candidate algorithms.
\end{itemize}
An example is the scenario of ``On-the-fly computing'' \cite{happe2013fly}, including ``On-the-fly machine learning'' \cite{mohrWTH19} as one of its instantiations, where users can request online (machine learning) software services customized towards their needs. Here, users are unwilling to wait for several hours until their service is ready, but rather claim a result quickly. Hence, for providing a first version of an appropriate service, costly search and online evaluations are not affordable. As will be seen, XAS offers a good compromise solution: Although it allows for the consideration of extremely many candidate solutions, and even offers the ability to recommend configurations that have never been encountered so far, it is still amenable to AS techniques and avoids costly online evaluations.
 


In a sense, XAS relates to standard AS as the emerging topic of extreme classification (XC) \cite{bengio2019extreme} relates to standard multi-class classification.
Similar to XC, the problem of learning from sparse data is a major challenge for XAS: For a single algorithm, there are typically only observations for a few instances.
In this paper, we propose a benchmark dataset for XAS and investigate the ability of state-of-the-art AS approaches to deal with this sparsity and to scale with the size of candidate sets.
Furthermore, to support more effective learning from sparse data, we propose methods based on ``dyadic'' feature representations, in which both problem instances and algorithms are represented in terms of feature vectors. In an extensive experimental study, we find these methods to yield significant improvements.

\section{From Standard to Extreme Algorithm Selection}
In the standard (per-instance) algorithm selection setting, first introduced in \cite{rice1976algorithm}, we are interested in finding a mapping $s:\mathcal{I} \longrightarrow \mathcal{A}$, called algorithm selector.
Given an instance $i$ from the instance space $\mathcal{I}$, the latter selects the algorithm $a^*$ from a set of candidate algorithms $\mathcal{A}$, optimizing a performance measure $m:\mathcal{I} \times \mathcal{A} \longrightarrow \mathbb{R}$. Furthermore, $m$ is usually costly to evaluate.
The optimal selector is called \emph{oracle} and is defined as
\begin{equation}\label{eq:oracle_definition}
    s^*(i) \defeq  \arg\max_{a \in \mathcal{A}} \mathbb{E}\big[ \, m(i,a) \, \big]
\end{equation} for all $i \in \mathcal{I}$. The expectation operator $\mathbb{E}$ accounts for any randomness in the application of the algorithm\,---\,in the non-deterministic case, the result of applying $a$ to $i$, and hence the values of the performance measure, are random variables.

Most AS approaches leverage machine learning techniques, in one way or another learning a surrogate (regression) model $\widehat{m}: \mathcal{I} \times \mathcal{A} \longrightarrow \mathbb{R}$, which is fast to evaluate and thus allows one to compute a selector $\widehat{s}: \mathcal{I} \longrightarrow \mathcal{A}$ by 
\begin{equation}\label{eq:surrogate_selector}
    \widehat{s}(i) \defeq \arg\max_{a \in \mathcal{A}} \widehat{m}(i,a) \, .
\end{equation}
In order to infer such a model, we usually assume the existence of a set of training instances $\mathcal{I}_D \subset \mathcal{I}$ for which we have instantaneous access to the associated performances of some or often all algorithms in $\mathcal{A}$ according to $m$.

The XAS setting distinguishes itself from the standard AS setting by two important properties. Firstly, we assume that the set of candidate algorithms $\mathcal{A}$ is \textit{extremely} large. Thus, approaches need to be able to scale well with the size of $\mathcal{A}$. Secondly, due to the size of $\mathcal{A}$, we can no longer reasonably assume to have evaluations for each algorithm on each training instance. Instead, we assume that the training matrix spanned by the training instances and algorithms is only sparsely filled. In fact, we might even have algorithms without any evaluations at all. Hence, suitable approaches need to be able to learn from very few data and to tackle the problem of ``zero-shot learning'' \cite{DBLP:journals/tist/WangZYM19}.

Similarly, the XAS setting differs from the AC and CASH settings in two main points. Firstly, dealing with real-valued hyperparameters, the set of (configured) algorithms $\mathcal{A}$ is generally assumed to be \textit{infinite} in both AC and CASH, whereas $\mathcal{A}$ is still finite (even if extremely large) in XAS. More importantly, in both AC and CASH, one usually assumes having time to perform online evaluations of solution candidates at recommendation time. However, as previously mentioned, this is not the case in XAS, where instantaneous recommendations are required. Hence, the XAS setting significantly differs from the AS, AC, and CASH settings. A summary of the main characteristics of these settings is provided in Table \ref{tab:setting-differences}.

\begin{table}[t]
    \centering
    \caption{Overview of the characteristics of the problem settings we distinguish.}
    \label{tab:setting-differences}
    \resizebox{.9\textwidth}{!}{
    \begin{tabular}{p{3.5cm}||c|c|c|c}
    \toprule
        Characteristics/Setting & AS & XAS & AC & CASH  \\
        \midrule
        Size of $\mathcal{A}$ & \, at most tens \, & \, extremely many \, & \, potentially infinite \, & \, potentially infinite \, \\
        Training data & complete & sparse & mostly not present & mostly not present \\
        Online evaluations & no & no & yes & yes \\
        \bottomrule
    \end{tabular}
    }
\end{table}

\section{Exploiting Instance Features}\label{sec:exploiting_instance_features}

Instance-specific AS is based on the assumption that instances can be represented in terms of feature information.
For this purpose, $f_I: \mathcal{I} \longrightarrow \mathbb{R}^k$ denotes a function representing instances as $k$-dimensional, real-valued feature vectors, which can be used to learn a surrogate model (\ref{eq:surrogate_selector}).
This can be done based on different types of data and using different loss functions.

\subsection{Regression}
\label{sec:reg1}

The most common approach is to tackle AS as a regression problem, i.e., to construct a regression dataset for each algorithm, where entries consist of an instance representation and the associated performance of the algorithm at question. Accordingly, the dataset associated with algorithm $a \in \mathcal{A}$ consists of tuples of the form $\big(f_I(i), m(i,a)\big)$, created for those instances $i \in \mathcal{I}_D$ to which $a$ has been applied, so that a performance evaluation $m(i,a) \in \mathbb{R}$ is available. Using this dataset, a standard regression model $\widehat{m}_a$ can be learned per algorithm $a$, and then used as a surrogate. The model can be realized as a neural network or a random forest, and trained on loss functions such as root mean squared or absolute error. For an overview of methods of this kind, we refer to Section~\ref{sec:related_work}.

This approach has two main disadvantages. Firstly, it is not well suited for the XAS setting, as it requires learning a huge number of surrogate models, one per algorithm. Although these models can usually be trained very quickly, the assumption of sparse training data in the XAS setting requires them to be learned from only a handful of training examples\,---\,it is not even uncommon to have algorithms without any performance value at all. Accordingly, the sparser the data, the more drastically this approach drops in performance, as will be seen in the evaluation in Section \ref{sec:experimental_evaluation}. Secondly, it requires precise real-valued evaluations of the measure $m$ as training information, which might be costly to obtain. In this regard, one may also wonder, whether regression is not solving an unnecessarily difficult problem: Eventually, AS is only interested in finding the best algorithm for a given problem instance, or, more generally, in ranking the candidate algorithms in decreasing order of their expected performance. An accurate prediction of absolute performances is a \emph{sufficient} but not a \emph{necessary} condition for doing so.

\subsection{Ranking}
\label{sec:ranking1}

As an alternative to regression, one may therefore think of tackling AS  as a \emph{ranking} problem. More specifically, the counterpart of the regression approach outlined above is called \emph{label ranking} (LR) in the literature \cite{vembu2010label}. Label ranking deals with learning to rank choice alternatives (referred to as ``labels'') based on given contexts represented by feature information. In the setting of AS, contexts and labels correspond to instances and algorithms, respectively. The type of training data assumed in LR consists of rankings $\pi_i$ associated with training instances $i \in \mathcal{I}_D$, that is, order relations of the form $(f_I(i), a_{i,1}) \succ \ldots \succ (f_I(i), a_{i,l_i})$, in which $\succ$ denotes an underlying preference relation; thus, $(f_I(i), a ) \succ (f_I(i), a' )$ means that, for instance $i$ represented by features $f_I(i)$, algorithm $a$ is preferred to (better than) algorithm $a'$. If $i$ is clear from the context, we also represent the ranking by $a_{1} \succ \ldots \succ a_{l_i}$. Compared to the case of regression, a ranking dataset of this form can be constructed more easily, as it only requires qualitative comparisons between algorithms instead of real-valued performance estimates.

A common approach to label ranking is based on the so-called Plackett-Luce (PL) model \cite{cheng2010label}, which specifies a parameterized probability distribution on rankings over labels (i.e., algorithms in our case). The underlying idea is to associate each algorithm $a$ with a latent utility function $\widehat{m}_a: \mathcal{I} \longrightarrow \mathbb{R}_+$ of a context (i.e., an instance), which estimates how well an algorithm is suited for a given instance. The functions $\widehat{m}_a$ are usually modeled as log-linear functions
\begin{equation}
    \widehat{m}_a(i) = \exp\big(\boldsymbol{\theta}_a^{\top} \, f_I(i)\big) \, ,
\end{equation} 
where $\boldsymbol{\theta}_a \in \mathbb{R}^k$ is a real-valued, $k$-dimensional vector, which has to be fit for each algorithm $a$. The PL model specifies a probability distribution on rankings: given an instance $i \in \mathcal{I}$, the probability of a ranking $a_1 \succ \ldots \succ a_z$ over any subset $\{ a_1, \ldots , a_z\} \subseteq \mathcal{A}$ is
\begin{equation}\label{eq:lr_probability}
    \mathbb{P}(a_1 \succ \ldots \succ a_z \, \vert\,  \boldsymbol{\Theta}) = \prod_{n=1}^{z} \frac{\widehat{m}_{a_n}(i)}{\widehat{m}_{a_n}(i) + \ldots + \widehat{m}_{a_z}(i)} \, .
\end{equation}
A probabilistic model of that kind suggests learning the parameter matrix $\boldsymbol{\Theta}= \{ \boldsymbol{\theta}_a \, | \, a \in \mathcal{A} \}$ via maximum likelihood estimation, i.e., by maximizing 
$$
L(\boldsymbol{\Theta}) = \prod_{i \in \mathcal{I}_D} \mathbb{P}(\pi_i \, \vert \, \boldsymbol{\Theta})
$$
associated with (\ref{eq:lr_probability}); this approach is explained in detail in \cite{cheng2010label}. Hence, the associated loss function under which we learn is now of a probabilistic nature (the logarithm of the PL-probability). It no longer focuses on the difference between the approximated performance $\widehat{m}_a(i)$ and the true performance $m(i,a)$, but on the ranking of the algorithms with respect to $m$\,---\,putting it in the jargon of preference learning, the former is a ``pointwise'' while the latter is a ``listwise'' method for learning to rank \cite{cao2007learning}. 

This approach potentially overcomes the second problem explained for the case of regression, but not the first one: It still fits a single model per algorithm $a$ (the parameter vector $\boldsymbol{\theta}_a$), which essentially disqualifies it for the XAS setting.

\subsection{Collaborative Filtering}\label{subsec:collaborative_filtering}
This may suggest yet another approach, namely the use of collaborative filtering (CF) \cite{goldberg1992using}, in the setting of AS originally proposed by \cite{stern2010collaborative}. In CF for AS, we assume a (usually sparse) performance matrix $R^{\vert \mathcal{I}_D \vert \times \vert \mathcal{A} \vert}$, where an entry $R_{i,a} = m(i,a)$ corresponds to the performance of algorithm $a$ on instance $i$ according to $m$ if known, and $R_{i,a}=\,?$ otherwise. CF methods were originally designed for large-scale settings, where products (e.g.\ movies) are recommended to users, and data to learn from is sparse. Hence, they appear to fit well for our XAS setting.

Similar to regression and ranking, model-based CF methods also learn a latent utility function. They do so by applying matrix factorization techniques to the performance matrix $R$, trying to decompose it into matrices $U \in \mathbb{R}^{\vert \mathcal{I}_D \vert \times t}$ and $V \in \mathbb{R}^{t \times \vert \mathcal{A} \vert}$ w.r.t.\ some loss function $L(R,U,V)$, such that 
\begin{equation}
    R \approx \widehat{R} = UV^{\top} \, ,
\end{equation} 
where $U$ ($V$) can be interpreted as latent features of the instances (algorithms), and $t$ is the number of latent features. Accordingly, the latent utility of a known algorithm $a$ for a known instance $i$ can be computed as
\begin{equation}
    \widehat{m}_a(i) = U_{i,\bullet} V_{\bullet,a}^{\top} \, ,
\end{equation} 
even if the associated value $R_{i,a}$ is unknown in the performance matrix used for training. The loss function $L(R,U,V)$ depends on the exact approach used\,---\,examples include the root mean squared error and the absolute error restricted by some regularization term to avoid overfitting. In \cite{misir2017alors}, the authors suggest a CF approach called Alors, which we will use in our experiments later on. It can deal with unknown instances by learning a feature map from the original instance to the latent instance feature space. Alors leverages the CF approach CoFi$^\text{RANK}$ \cite{weimer2008cofi} using the normalized discounted cumulative gain (NDCG) \cite{wang2013theoretical} as loss function $L(R,U,V)$. Since the NDCG is a ranking loss, it focuses on decomposing the matrix $R$ so as to produce an accurate ranking of the algorithms. More precisely, it uses an exponentially decaying weight function for ranks, such that more emphasis is put on the top and less on the bottom ranks. Hence, it seems particularly well suited for our use case.

\section{Dyadic Feature Representation}

As discussed earlier, by leveraging instance features, or learning such a representation as in the case of Alors, the approaches presented in the previous section can generalize over instances. Yet, none of them scales well to the XAS setting, as they do not generalize over algorithms; instead, the models are algorithm-specific and trained independently of each other. For the approaches presented earlier (except for Alors), this does not only result in a large number of models but also requires these models to be trained on very few data. Furthermore, it is not uncommon to have algorithms without any observation.
A natural idea, therefore, is to leverage feature information on algorithms as well. 

More specifically, we use a feature function $f_A: \mathcal{A} \longrightarrow \mathbb{R}^d$ representing algorithms as $d$-dimensional, real-valued feature vectors. Then, instead of learning one latent utility model per algorithm, the joint feature representation of a ``dyad'' consisting of an instance and an algorithm, allows us to learn a single joint model 
\begin{equation}\label{eq:dyadic_joint_model}
     \widehat{m}: f_I(\mathcal{I}) \times f_A(\mathcal{A}) \longrightarrow \mathbb{R} \, ,
\end{equation}
and hence to estimate the performance of a given algorithm $a$ on a given instance $i$ in terms of $\widehat{m}(f_I(i), f_A(a))$. 

\subsection{Regression}\label{subsec:dyad_feature_regression}

With the additional feature information at hand, instead of constructing one dataset per algorithm, we resolve to a single joint dataset comprised of examples $\Big(\psi\big(f_I(i),f_A(a)\big), m(i,a)\Big)$ with dyadic feature information for all instances $i \in \mathcal{I}_D$ and algorithms $a \in \mathcal{A}$ for which a performance value $m(i,a)$ is known. Here, 
\begin{equation}\label{eq:jfm}
\psi: \mathbb{R}^k \times \mathbb{R}^d \longrightarrow \mathbb{R}^q
\end{equation}
is a joint feature map that defines how the instance and algorithm features are combined into a single feature representation of a dyad. 
What is sought, then, is a (parametrized) latent utility function $\widehat{m}_{\boldsymbol{\theta}}: \mathbb{R}^q \longrightarrow \mathbb{R}$, such that
\begin{equation}\label{eq:jointmodel}
 \widehat{m}_{\boldsymbol{\theta}}\Big( \psi \big(f_I(i),f_A(a) \big) \Big) \,
\end{equation}
is an estimation of the performance of algorithm $a$ on instance $i$.
Obviously, the choice of  $\psi$ will have an important influence on the difficulty of the regression problem and the quality of the model (\ref{eq:jointmodel}) induced from the data $\mathcal{D}^{\mathit{REG}}$. The regression task itself comes down to learning the parameter vector $\boldsymbol{\theta}$. In principle, this can be done exactly as in Section \ref{sec:reg1}, also using the same loss function. Note that this is a generalization of the approach used by SMAC \cite{hutterHL11} for predicting performances across instances in algorithm configuration. We allow for a generic joint feature map $\psi$ and an arbitrary model for $\widehat{m}_{\boldsymbol{\theta}}$, whereas SMAC limits itself to a concatenation of features and trains a random forest for modeling $\widehat{m}_{\boldsymbol{\theta}}$. Once again, it is noteworthy that SMAC by itself is not applicable in the XAS setting, as it relies on costly online evaluations.

\subsection{Ranking}

A similar adaptation can be made for the (label) ranking approach presented in Section \ref{sec:ranking1} \cite{tornede2019algorithm}. Formally, this corresponds to a transition from the setting of label ranking to the setting of \emph{dyad ranking} (DR) as recently proposed in \cite{schaefer2018dyad}. 
The first major change in comparison to the setting of label ranking concerns the training data, where the rankings
$\pi_i$ over subsets of algorithms $\{ a_{i,1}, \ldots , a_{i,l_i}\} \subseteq \mathcal{A}$ for instance $i$ are now of the form
\begin{equation}
    \psi\big(f_I(i), f_A(a_{i,1})\big) \succ \ldots \succ \psi\big(f_I(i), f_A(a_{i,l_i})\big) \, .
\end{equation}
Thus, we no longer represent an algorithm $a$ simply by its label ($a$) but by features $f_A(a)$. Furthermore, like in the case of regression, we no longer learn one latent utility function per algorithm, but a single model of the form (\ref{eq:jointmodel}) based on a dyadic feature representation. In particular, we model $\widehat{m}_{\boldsymbol{\theta}}$ as a feed-forward neural network, where $\boldsymbol{\theta}$ represents its weights, which, as shown in \cite{schaefer2018dyad}, can be learned via maximum likelihood estimation on the likelihood function implied by the underlying PL model. Note that the use of a neural network is of particular interest here, since it allows one to learn the underlying joint feature map $\phi$ implicitly. Although both instance and algorithm features are simply fed as a concatenated vector into the network, it can recombine these features due to its structure and thus implicitly learn such a joint feature representation.

In contrast to the methods presented in the previous section, the methods based on dyadic feature information are capable of assigning a utility to unknown algorithms. Thus, they are well suited for the XAS setting and in principle even applicable when $\mathcal{A}$ is infinite, as long as a suitable feature representation $f_A$ is available. Furthermore, as demonstrated empirically in Section \ref{sec:experimental_evaluation}, the dyadic feature approaches are very well suited for dealing with sparse performance matrices that are typical of the XAS setting.

\newcommand{\pareg}{\texttt{PAReg}}
\newcommand{\A}{\mathcal{A}}
\newcommand{\I}{\mathcal{I}}
\newcommand{\m}{$m$}
\section{Experimental Evaluation}\label{sec:experimental_evaluation}
In our experiments, we evaluate well established state-of-the-art approaches to algorithm selection as well as the proposed dyadic approaches in the XAS setting.
More specifically, we consider the problem of selecting a machine learning classifier (algorithm) for a new classification dataset (instance) as a case study related to the ``on-the-fly machine learning'' scenario \cite{mohrWTH19}. Please note that this is just one amongst many conceivable instantiations of the XAS setting, which is supposed to demonstrate the performance of the presented methods. To this end, we first generate a benchmark and then use this benchmark for comparison.
The generated benchmark dataset as well as the implementation of the approaches including detailed documentation is provided on GitHub\footnote{\url{https://github.com/alexandertornede/extreme_algorithm_selection}}. 

\subsection{Benchmark Dataset}
\label{ssec:benchmark}
In order to benchmark the generalization performance of the approaches presented above in the XAS setting, we consider the domain of machine learning. More precisely, the task is to select a classification algorithm for an (unseen) dataset. Therefore, a finite set of algorithms $\A$ for classification and a set of instances $\I$ corresponding to classification datasets need to be specified. Furthermore, a performance measure \m\ is needed to score the algorithms' performance.

The set of candidate algorithms $\A$ is defined by sampling up to 100 different parameterizations of 18 classification algorithms from the machine learning library WEKA \cite{frank2016weka}, ensuring these parameterizations not being too similar.
An overview of the algorithms, their parameters and the number of instantiations contained in $\mathcal{A}$ is given in Table~\ref{tbl:learners}.
This yields $|\A| = 1,270$ algorithms in total.
The last row of the table sums up the items of the respective column, providing insights into the dimensionality of the space of potential candidate algorithms.
    
The set of instances $\mathcal{I}$ is taken from the OpenML CC-18 benchmarking suite\footnote{\url{https://docs.openml.org/benchmark/\#openml-cc18} (Excluding datasets 554, 40923, 40927, 40996 due to technical issues.)} \cite{vanshoren2013openml}, which is a curated collection of various classification datasets that are considered interesting from a model selection resp.~hyperparameter optimization point of view.
This property makes the datasets particularly appealing for the XAS benchmark dataset, as it ensures more diversity across the algorithms.

Accordingly, the total performance matrix spanned by the algorithms and classification datasets in principle features $1,270 \cdot 71 = 88,900$ entries for which the benchmark contains $55,919$ actual values and the rest are unknown.
    
In the domain of machine learning, one is usually more interested in the generalization performance of an algorithm than in the runtime. Therefore, $m$ is chosen to assess the solution performance of an algorithm.
To this end, we carry out a 5-fold cross validation and measure the mean accuracy across the folds\footnote{The standard deviation of the performance values per dataset is on average $0.101$, minimum $0.0064$ and maximum $0.33$.}.
As the measure of interest, accuracy is a reasonable though to some extent arbitrary choice. Note that in principle any other measure could have been used for generating the benchmark as well.


\begin{table}[t]
    \centering
    \caption{The table shows the types of classifiers used to derive the set $\A$. Additionally, the number of numerical parameters (\#num.P), categorical parameters (\#cat.P), and instantiations (n) is shown.}
    \label{tbl:learners}
    \resizebox{.65\columnwidth}{!}{
\begin{tabular}[h]{crrrrrrrrrrrrrrrrrr}
    \toprule
     Learner & \rotatebox{90}{0R}& \rotatebox{90}{1R}& \rotatebox{90}{BN}& \rotatebox{90}{DS}& \rotatebox{90}{DT}& \rotatebox{90}{IBk}& \rotatebox{90}{J48}& \rotatebox{90}{JR}& \rotatebox{90}{KS}& \rotatebox{90}{L}& \rotatebox{90}{LMT}& \rotatebox{90}{MP}& \rotatebox{90}{NB}& \rotatebox{90}{PART}& \rotatebox{90}{REPT}& \rotatebox{90}{RF}& \rotatebox{90}{RT}& \rotatebox{90}{SMO} \\
     \midrule
     \#num.P& 0&1&0&0&1&1&2&2&1&1&2&2&0&2&3&3&4&1\\ 
     \#cat.P& 0 &0 &2 &0 &3 &3 &6 &2 &2 &0 &5 &6 &2 &2& 2& 2& 4& 2\\ 
     n & 1 & 30& 12& 1 &45& 89& 100& 100& 99& 100& 100& 100 & 3 &91& 100& 99& 100& 100\\
     \bottomrule
\end{tabular}
    }
\end{table}

Training data for CF and regression-based approaches can then be obtained by using the performance values as labels.
In contrast, for training ranking approaches, the data is labeled with rankings derived by ordering the algorithms in a descending order w.r.t.~their performance values.
Note that information about the exact performance value itself is lost in ranking approaches.  

We would like to note that the problem underlying this benchmark dataset could of course be cast as an AC or CASH problem. However, here we make the assumption that there is no time for costly online evaluations due to the on-the-fly setting and hence standard AC and CASH methods are not applicable.

\medskip

\noindent
\textbf{Instance Features.}
For the setting of machine learning, the instances are classification datasets and associated feature representations are called meta-features \cite{nguyen2014using}.
To derive a feature description of the datasets, we make use of a specific subclass of meta-features called \textit{landmarkers}, which are performance scores of cheap-to-validate algorithms on the respective dataset.
More specifically, we use all 45 landmarkers as provided by OpenML \cite{vanshoren2013openml}, for which different configurations of the following learning algorithms are evaluated based on the error rate, area under the (ROC) curve, and Kappa coefficient: Naive Bayes, One-Nearest Neighbour, Decision Stump, Random Tree, REPTree and J48. Hence, in total we use 45 features to represent a classification dataset.
%
%

\medskip

\noindent
\textbf{Algorithm Features.}
The presumably most straight-forward way of representing an algorithm in terms of a feature vector is to use the values of its hyperparameters.
Thus, we can describe each individual algorithm by a vector of their hyperparameter-values.
Based on this, the general feature description is obtained by concatenation of the vectors. As already mentioned, the neural network-based dyad ranking approach implicitly learns a more sophisticated joint feature map.
Due to the way in which we generated the set of candidate algorithms $\A$, we can compress the vector sharing features for algorithms of the same type. Additionally, we augment the vector by a single categorical feature denoting the type of algorithm.
Given any candidate algorithm, its feature representation is obtained by setting the type of algorithm indicator feature to its type, each element of the vector corresponding to one of its hyperparameters to the specific value, and other entries to $0$. Categorical parameteters, i.e. features, are one-hot encoded yielding a total of 153 features to represent an algorithm.



\subsection{Baselines}
\label{ssec:baselines}
To better relate the performance of the different approaches to each other and to the problem itself, we employ various baselines. While \randomRank\ assigns ranks to algorithms simply at random, \avgPrf\ first averages the observed performance values for each candidate algorithm and predicts the ranking according to these average performances. $k$-NN LR retrieves the $k$ nearest neighbors from the training data, averages the performances and predicts the ranking which is induced by the average performances. Since \avgRank\ is commonly used as another baseline in the standard AS setting, we note that we omit this baseline on purpose. This is because meaningful average ranks of algorithms are difficult to compute in the XAS setting, where the number of algorithms evaluated, and hence the length of the rankings of algorithms, vary from  dataset to dataset.

\subsection{Experimental Setup}
\label{ssec:setup}

In the following experiments, we investigate the performance of the different approaches and baselines in the setting of XAS for the example of the proposed benchmark dataset as described in Section~\ref{ssec:benchmark}.

We conduct a 10-fold cross validation to divide the dataset into 9 folds of known and 1 fold of unknown instances.
From the resulting set of known performance values, we then draw a sample of 25, 50, or 125 pairs of algorithms for every instance under the constraint that the performances of the two algorithms is not identical.
Thus, a maximum fill degree of 4\%, 8\% respectively 20\% of the performance matrix is used for training, as algorithms may occur more than once in the sampled pairs.
The sparse number of training examples is motivated by the large number of algorithms in the XAS setting. The assumption that performance values are only available for a small subset of the  algorithms is clearly plausible here.
Throughout the experiments, we ensure that all approaches are provided the same instances for training and testing, and that the label information is at least based on the same performance values.

In the experiments, we compare various models with each other.
This includes two versions of \texttt{Alors}, namely \AlorsRegr\ and \AlorsNDCG\ optimizing for a regression respectively ranking loss.
Furthermore, we consider a state-of-the-art regression approach learning a RandomForest regression model per algorithm (\pareg). Note that for those algorithms with no training data at all, we make \pareg\  predict a performance of $0$, as recommending such an algorithm does not seem reasonable. 
Lastly, we consider two approaches leveraging a dyadic feature representation, internally fitting either a RandomForest for regression (\texttt{DFReg}) or a feed-forward neural network for ranking (\texttt{DR}). For both dyadic approaches, the simple concatenation of instance and algorithm features is used as a feature map.
In contrast to the other methods, the ranking model is only provided the information which algorithm of a sampled pair performs better, as opposed to the exact performance value that is given to other methods.
A summary of the type of features and label information used by the different approaches/baselines is given on the left side of Table~\ref{tab:approach-details}.

\begin{table}[t]
    \centering
    \caption{Overview of the data provided to the approaches and their applicability to the considered scenarios.}
    \label{tab:approach-details}
    \resizebox{.75\textwidth}{!}{
    \begin{tabular}{p{.6cm}p{3cm}ccc||p{.6cm}p{3cm}cccccc}
        \toprule
         & Approach & $f_I$ & $f_A$ & Label & & Approach & $f_I$ & $f_A$ & Label \\
        \midrule
        \multirow{5}{*}{\rotatebox{90}{approaches}} &
          \texttt{Alors (REGR)} & \cmark & \xmark & $m$ & \multirow{4}{*}{\rotatebox{90}{baselines}}&\texttt{RandomRank} & \xmark & \xmark & \\
         &\texttt{Alors (NDCG)} & \cmark & \xmark & $m$ & & \texttt{AvgPrfm} & \xmark & \xmark & $m$\\
         &\texttt{PAReg} & \cmark & \xmark & $m$ & & \texttt{AvgRank} & \xmark & \xmark & $\pi$\\
         &\texttt{DFReg} & \cmark & \cmark & $m$ & & \texttt{$k$-NN LR} & \cmark & \xmark & $m$\\
         &\texttt{DR} & \cmark & \cmark & $\pi$  \\
         \bottomrule
    \end{tabular}
    }
\end{table}

The test performance of the approaches is evaluated by sampling $10$ algorithms for every (unknown) instance to test for.
The comparison is done with respect to different metrics detailed further below, and the outlined sampling evaluation routine is repeated $100$ times.

Statistical significance w.r.t~performance differences between the best method and any other method is determined by a Wilcoxon rank sum test with a threshold of 0.05 for the p-value. Significant improvements of the best method over another one is indicated by $\bullet$.

Experiments were run on nodes with two Intel Xeon Gold ``Skylake'' 6148 with 20 cores each and 192 GB RAM.
\subsection{Performance Metrics}
On the test data, we compute the following performance metrics measuring desirable properties of XAS approaches.

\textbf{\regretAtK{k}} is the difference between the performance value of the best algorithm within the predicted top-$k$ of algorithms and the actual best algorithm.
The domain of \regretAtK{k} is $[0,1]$, where 0 is the optimum meaning no regret.

\textbf{NDCG$@k$} is a position-dependent ranking measure (\textit{n}ormalized \textit{d}iscounted \textit{c}umulative \textit{g}ain) to measure how well the ranking of the top-$k$ algorithms can be predicted. It is defined as
\begin{equation*}
    \text{NDCG}@k(\pi, \pi^*) = \frac{\text{DCG}@k(\pi)}{\text{DCG}@k(\pi^*)} = \dfrac{\left(\sum\limits_{n=1}^k \frac{2^{m(i,\pi_n) - 1}}{\log(n + 2)}\right)}{\left(\sum\limits_{n=1}^k \frac{2^{m(i,\pi^*_n) - 1}}{\log(n + 2)}\right)}\, ,
\end{equation*}
where $i$ is a (fixed) instance, $\pi$ is a ranking and $\pi^*$ the optimal ranking, and $\pi_n$ gives the algorithm on rank $n$ in ranking $\pi$.
The NDCG emphasizes correctly assigned ranks at higher positions with an exponentially decaying importance.
NDCG ranges in $[0,1]$, where $1$ is the optimal value.

\textbf{Kendall's $\tau$} is a rank correlation measure. Given two rankings (over the same set of elements) $\pi$ and $\pi^\prime$, it is defined as
\begin{equation}
    \tau(\pi,\pi^\prime) = \frac{C - D}{\sqrt{(C + D + T_{\pi}) \cdot (C + D + T_{\pi^\prime})}}
\end{equation}
where $C$/$D$ is the number of so-called \textit{concordant}/\textit{discordant} pairs in the two rankings, and $T_{\pi}$/$T_{\pi^\prime}$ is the number of ties in $\pi$/$\pi^\prime$.
Two elements are called a concordant/discordant pair if their order within the two rankings is identical/different, and tied if they are on the same rank.
Intuitively, this measure determines on how many pairs the two rankings coincide.
It takes values in $[-1,1]$, where $0$ means uncorrelated, $-1$ inversely, and $1$ perfectly correlated.

\subsection{Results}
\label{ssec:results}

The results of the experiments are shown in Table~\ref{tbl:results}.
It is clear from the table that the methods for standard algorithm selection tend to fail especially in the scenarios with only few algorithm performance values per instance.
This includes the approach of building a distinct regression model for each algorithm (\texttt{PAReg}) as well as for the collaborative filtering approach \texttt{Alors}, independently of the loss optimized for, even though the NDCG variant has a slight edge over the regression one.
Moreover, \texttt{Alors} even fails to improve over simple baselines, such as \texttt{AvgPerformance} and \texttt{$k$-NN LR}.
With an increasing number of training examples, \texttt{PAReg} improves over the baselines and also performs better than \texttt{Alors}, but never yields the best performance for any of the considered settings or metrics.

\begin{table}[t]
    \centering
    \caption{Results for the performance metrics Kendall' tau ($\tau$), NDCG@k (N@3, N@5), and \regretAtK{k} (R@1, R@3) for varying number of performance value pairs used for training.
    The best performing approach is highlighted in bold, the second best is underlined, and significant improvements of the best approach over others is denoted by $\bullet$.
    }
    \label{tbl:results}
    \resizebox{\textwidth}{!}{
        
\begin{tabular}{p{2.6cm} |  lllll | lllll}
\toprule
\multirow{2}{*}{Approach} & \multicolumn{5}{c|}{4\% fill rate / 25 performance value pairs} & \multicolumn{5}{c}{8\% fill rate / 50 performance value pairs} \\
&$\tau$&N@3&N@5&R@1&R@3&$\tau$&N@3&N@5&R@1&R@3\\
\midrule

\texttt{PAReg} & 0.1712 $\bullet$ & 0.9352 $\bullet$ & 0.9433 $\bullet$ & 0.0601 $\bullet$ & 0.0185 $\bullet$ 
& 0.2537 $\bullet$ & 0.9453 $\phantom{\circ}$ & 0.9594 $\phantom{\circ}$ & 0.0493 $\phantom{\circ}$ & 0.0136 $\phantom{\circ}$\\

\texttt{Alors (NDCG)} & 0.0504 $\bullet$ & 0.9205 $\bullet$ & 0.9223 $\bullet$ & 0.0686 $\bullet$ & 0.0225 $\phantom{\circ}$ 
& 0.0472 $\bullet$ & 0.9155 $\bullet$ & 0.9164 $\bullet$ & 0.0614 $\bullet$ & 0.0208 $\phantom{\circ}$\\
\texttt{Alors (REGR)} & 0.0303 $\bullet$ & 0.9117 $\bullet$ & 0.9191 $\bullet$ & 0.0794 $\bullet$ & 0.0190 $\bullet$ 
& 0.0807 $\bullet$ & 0.9172 $\bullet$ & 0.9304 $\bullet$ & 0.0754 $\bullet$ & 0.0285 $\bullet$\\

\texttt{DR} & \underline{0.3445} $\phantom{\circ}$ & \underline{0.9523} $\phantom{\circ}$ & \underline{0.9604} $\phantom{\circ}$ & 0.0381 $\phantom{\circ}$ & \underline{0.0089} $\phantom{\circ}$ 
& \textbf{0.3950} $\phantom{\circ}$ & \textbf{0.9584} $\phantom{\circ}$ & \textbf{0.9685} $\phantom{\circ}$ & \underline{0.0322} $\phantom{\circ}$ & \textbf{0.0087} $\phantom{\circ}$\\

\texttt{DFReg} & \textbf{0.3819} $\phantom{\circ}$ & \textbf{0.9564} $\phantom{\circ}$ & \textbf{0.9652} $\phantom{\circ}$ & \textbf{0.0302} $\phantom{\circ}$ & \textbf{0.0079} $\phantom{\circ}$ 
& \underline{0.3692} $\phantom{\circ}$ & \underline{0.9573} $\phantom{\circ}$ & \underline{0.9661} $\phantom{\circ}$ & \textbf{0.0300} $\phantom{\circ}$ & \underline{0.0123} $\phantom{\circ}$\\
%
\midrule

\texttt{RandomRank} & -0.0038 $\bullet$ & 0.8933 $\bullet$ & 0.9105 $\bullet$ & 0.0878 $\bullet$ & 0.0272 $\bullet$ 
& -0.0038 $\bullet$ & 0.8933 $\bullet$ & 0.9105 $\bullet$ & 0.0878 $\bullet$ & 0.0272 $\bullet$\\

\texttt{AvgPerformance} & 0.1384 $\bullet$ & 0.9388 $\bullet$ & 0.9433 $\bullet$ & \underline{0.0337} $\phantom{\circ}$ & 0.0090 $\phantom{\circ}$ 
& 0.2083 $\bullet$ & 0.9355 $\bullet$ & 0.9508 $\bullet$ & 0.0493 $\bullet$ & 0.0199 $\bullet$\\

\texttt{1-NN LR} & 0.1227 $\bullet$ & 0.9290 $\bullet$ & 0.9310 $\bullet$ & 0.0733 $\bullet$ & 0.0230 $\bullet$ 
& 0.1059 $\bullet$ & 0.9246 $\bullet$ & 0.9296 $\bullet$ & 0.0564 $\bullet$ & 0.0209 $\phantom{\circ}$\\

\texttt{2-NN LR} & 0.1303 $\bullet$ & 0.9278 $\bullet$ & 0.9310 $\bullet$ & 0.0642 $\bullet$ & 0.0193 $\bullet$ 
& 0.0874 $\bullet$ & 0.9269 $\bullet$ & 0.9343 $\bullet$ & 0.0541 $\bullet$ & 0.0206 $\phantom{\circ}$\\

\midrule
\multicolumn{3}{c}{ } \\











\cmidrule{1-6}
\multirow{2}{*}{Approach} & \multicolumn{5}{c|}{20\% fill rate / 125 performance value pairs} \\
&$\tau$&N@3&N@5&R@1&R@3\\
\cmidrule{1-6}

\texttt{PAReg} & 0.3003 $\bullet$ & 0.9525 $\phantom{\circ}$ & 0.9632 $\phantom{\circ}$ & 0.0395 $\phantom{\circ}$ & 0.0107 $\phantom{\circ}$\\

\texttt{Alors (NDCG)} & 0.0540 $\bullet$ & 0.9220 $\bullet$ & 0.9242 $\bullet$ & 0.0542 $\bullet$ & 0.0228 $\bullet$\\
\texttt{Alors (REGR)} & 0.1039 $\bullet$ & 0.9160 $\bullet$ & 0.9329 $\bullet$ & 0.0604 $\bullet$ & 0.0222 $\bullet$\\

\texttt{DR} & \textbf{0.4507} $\phantom{\circ}$ & \textbf{0.9696} $\phantom{\circ}$ & \underline{0.9715} $\phantom{\circ}$ & \textbf{0.0241} $\phantom{\circ}$ & \textbf{0.0055} $\phantom{\circ}$\\

\texttt{DFReg} & \underline{0.4264} $\phantom{\circ}$ & \underline{0.9629} $\phantom{\circ}$ & \textbf{0.9720} $\phantom{\circ}$ & \underline{0.0292} $\phantom{\circ}$ & \underline{0.0071} $\phantom{\circ}$\\
%
\cmidrule{1-6}

\texttt{RandomRank} & -0.0038 $\bullet$ & 0.8933 $\bullet$ & 0.9105 $\bullet$ & 0.0878 $\bullet$ & 0.0272 $\bullet$\\

\texttt{AvgPerformance} & 0.2541 $\bullet$ & 0.9437 $\bullet$ & 0.9536 $\bullet$ & 0.0523 $\bullet$ & 0.0084 $\phantom{\circ}$\\

\texttt{1-NN LR} & 0.1152 $\bullet$ & 0.9245 $\bullet$ & 0.9318 $\bullet$ & 0.0594 $\bullet$ & 0.0249 $\bullet$\\

\texttt{2-NN LR} & 0.1142 $\bullet$ & 0.9292 $\bullet$ & 0.9350 $\bullet$ & 0.0412 $\phantom{\circ}$ & 0.0176 $\bullet$\\
\cline{1-6}
\end{tabular}

    }
\end{table}

In contrast to this, the proposed dyadic feature approaches clearly improve over both the methods for the standard AS setting and the considered baselines for all the metrics. 
Interestingly, \texttt{DFReg} performs best for the setting with only 25 performance value pairs, while \texttt{DR} has an edge over \texttt{DFReg} for the other two settings.
Still, the differences between the dyadic feature approaches are never significant, whereas  significant improvements can be achieved in comparison to the baselines and the other AS approaches.

Moreover, our study demonstrates the heterogeneity of the benchmark dataset. As described in \cite{schneiderH12}, a relevant measure for heterogeneity is the per-instance potential for improvement over a solution that is static across instances, i.e., what is often called the single best algorithm or solver (SBS). In this case study, the SBS is represented by the \avgPrf{} baseline, which is always worse than the oracle with respect to all measures and in particular the \regretAtK{k} measures. Hence, as the superior performances of our approach compared to the \avgPrf{} demonstrate, the benchmark dataset offers a potential for per-instance algorithm selection.

The results of our study show that models with strong generalization performance can be obtained despite the small number of training examples. Moreover, the results suggest that there is a need for the development of specific methods addressing the characteristics of the XAS setting.
This concerns the large number of different candidate algorithms as well as the sparsity of the training data.

\section{Related Work}\label{sec:related_work}
As most closely related work, we subsequently highlight several AS approaches to learning latent utility functions. For an up-to-date survey, we refer to \cite{kerschke2019automated}.

A prominent example of a method learning a regression-based latent utility function is \cite{xu2008satzilla}, which features an empirical hardness model per algorithm for estimating the runtime of an algorithm, i.e., its performance, for a given instance based on a ridge regression approach in the setting of SAT solver selection.
Similarly, \cite{leyton2002learning} learn per-algorithm hardness models using statistical (non-)linear regression models for algorithms solving the winner determination problem.
Depending on whether a given SAT instance is presumably satisfiable or not, conditional runtime prediction models are learned in \cite{haim2009restart} using ridge linear regression.

In \cite{cunha2018label}, a label-ranking-based AS approach for selecting collaborative filtering algorithms in the context of recommender systems is presented leveraging nearest neighbor and random forest label rankers.

Similar to our work, \cite{oentaryoHL15} leverages algorithm features in the form of a binary vector indicating which algorithm is considered to learn a probabilistic ranking model considering up to tens of algorithms.
AS was first modeled as a CF problem in \cite{stern2010collaborative}, using a probabilistic matrix factorization technique to select algorithms for the constraint solving problem.
Assuming a complete performance matrix for training, low-rank latent factors are learned in \cite{malitsky2014latent} using singular value decomposition to obtain a selector en par with the oracle.
Lastly, in \cite{tornede2020run2survive} a decision-theoretic approach is proposed leveraging survival analysis to explicitly acknowledge timeouts of algorithms in the learning process.


\section{Conclusion}
In this paper, we introduced the extreme algorithm selection (XAS) setting and investigated the scalability of various algorithm selection approaches in this setting.
To this end, we defined a benchmark based on the OpenML CC-18 benchmark suite for classification and a set of more than 1,200 candidate algorithms.
Furthermore, we proposed the use of dyadic approaches, specifically dyad ranking, taking into account feature representations of both problem instances (datasets) and algorithms, which allows them to work on very few training data.
In an extensive evaluation, we found that the approaches exploiting dyadic feature representations perform particularly well according to various metrics on the proposed benchmark and outperform other state-of-the-art AS approaches developed for the standard AS setting.

The currently employed algorithm features allow for solving the cold start problem only to a limited extent, i.e., only algorithms featuring known hyperparameters can be considered as new candidate algorithms.
Investigating features to describe completely new algorithms is a key requirement for the approaches considered in this paper, and therefore an important direction for future work.

\section*{Acknowledgements}
This work was supported by the German Research Foundation (DFG) within the Collaborative Research Center ``On-The-Fly Computing'' (SFB 901/3 project no.\ 160364472) and by the Paderborn Center for Parallel Computing (PC$^2$) through computing time.
%
%
%
\bibliographystyle{splncs04}
\bibliography{ijcai20}
\end{document}